\documentclass[conference]{IEEEtran}
\IEEEoverridecommandlockouts
\usepackage{cite}
\usepackage{amsmath,amssymb,amsfonts}
\usepackage{algorithmic}
\usepackage{graphicx}
\usepackage{textcomp}
\usepackage{xcolor}
\usepackage{tabularx}

\def\BibTeX{{\rm B\kern-.05em{\sc i\kern-.025em b}\kern-.08em
    T\kern-.1667em\lower.7ex\hbox{E}\kern-.125emX}}
    
\usepackage{fancyhdr}
\begin{document}

\title{The Invariant Ground Truth of Affect\\
\thanks{This work has been supported by the EU Horizon 2020 research and innovation programme from the TAMED project (GA No. 101003397).}
}


\author{\IEEEauthorblockN{
Konstantinos Makantasis,
Kosmas Pinitas,
Antonios Liapis,
Georgios N. Yannakakis\\
\IEEEauthorblockA{Institute of Digital Games, University of Malta, Msida, Malta.\\
Email: \{konstantinos.makantasis, kosmas.pinitas, antonios.liapis, georgios.yannakakis\}@um.edu.mt}}}

\maketitle
\thispagestyle{fancy}

\begin{abstract}
Affective computing strives to unveil the unknown relationship between affect elicitation, manifestation of affect and affect annotations. The ground truth of affect, however, is predominately attributed to the affect labels which inadvertently include biases inherent to the subjective nature of emotion and its labeling. The response to such limitations is usually augmenting the dataset with more annotations per data point; however, this is not possible when we are interested in self-reports via first-person annotation. Moreover, outlier detection methods based on inter-annotator agreement only consider the annotations themselves and ignore the context and the corresponding affect manifestation. This paper reframes the ways one may obtain a reliable ground truth of affect by transferring aspects of causation theory to affective computing. In particular, we assume that the ground truth of affect can be found in the causal relationships between elicitation, manifestation and annotation that remain \emph{invariant} across tasks and participants. To test our assumption we employ causation inspired methods for detecting outliers in affective corpora and building affect models that are robust across participants and tasks. We validate our methodology within the domain of digital games, with experimental results showing that it can successfully detect outliers and boost the accuracy of affect models. To the best of our knowledge, this study presents the first attempt to integrate causation tools in affective computing, making a crucial and decisive step towards general affect modeling.
\end{abstract}


\section{Introduction}
Affect modeling is largely viewed as a supervised learning task of subjectively defined labels. Due to labels' subjective nature, however, the ground truth of affect (which is attributed to such labels) is not easy to obtain reliably \cite{yannakakis2018ordinal}. The dominant approach for creating consistent affective corpora is to treat the presence of subjectivity as a nuisance and to attempt to engineer it away by collecting ever larger datasets \cite{mollahosseini2017affectnet} or employing several independent annotators \cite{artstein2017inter}. Although larger datasets may boost the performance of affect models and inter-annotator agreement methods can yield ``clean'' (i.e. consistent or reliable) labels, arguably both approaches aim to \emph{objectify} the labels of emotions, which are by nature subjective.

In this study we embrace subjectivity as an inherent property of affect labels and introduce the notion of \emph{causality} as a necessary element for reliable affect modeling. Inspired by causation theory \cite{scholkopf2021toward} we assume that there exist features of affect manifestation that are \emph{invariantly} related with affect labels. This implies that the relation that maps a subset of input features to affect labels remains invariant across different users or tasks. We name the subset of features for which such a relation exists as \textit{invariant features}. Our assumption is not new; on the contrary, every methodology for constructing handcrafted features within Affective Computing (AC) aims to transform affect manifestations to invariant features by implicitly assuming that such features exist. It is important to note that the notion of invariance we investigate in this paper is closely related to causation theory \cite{peters2017elements} where the existence of invariant features is a necessary---yet not sufficient---condition for the discovery of cause-effect relations.  


To validate the aforementioned assumption, we introduce a method for automatically identifying the set of invariant features and use it to detect outliers; i.e. data points stemming from a specific annotator or task for which the relation that maps the invariant features to affect is different than the majority of data. Specifically, we consider as invariant those features which present similar correlation patterns with affect labels across different annotators and tasks. Thus, we create representations based on the computed correlation patterns, which reveal to which degree the data from different annotators or tasks exhibit a consistent relation between input features and affect labels. Finally, those representations serve as input to a conventional outlier detection algorithm, which brings together annotators and tasks that exhibit similar correlation patterns between input features and affect (defined as inliers), and, at the same time, identifies and discards data from outliers. After the outliers' removal, we use the identified invariant feature set to build robust models of affect. 


Our method is of particular importance for AC corpora where disjoint sets of the data have been annotated by different annotators, such as first-person annotations. For this reason, we test our methodology within the domain of digital games using the first-person annotated AGAIN dataset \cite{melhart2022again}. Specifically we use data from fifty different participants that played three different games and self-annotated their gameplay in terms of arousal. Experimental results indicate that our method can effectively detect and remove outliers---i.e. inconsistent participants---resulting in robust models of affect. The boost in performance before and after outliers' removal is over $14\%$ for one out of the three games and over $6.5\%$ for the other two games. Notably, the performance of affect modeling for the set of outliers is on par with a random predictor.

This paper is novel in a number of ways. First, to the best of our knowledge, this study presents the first attempt to integrate causation tools in affect modeling; causation is expected to boost the generalization ability of affect models and arguably, generalizability is of utmost importance for any affective modeling method. Second, unlike any outlier detection method existent in AC, this paper proposes a methodology that detects outliers by considering both the affect measurements---i.e. the model's input---and the subjectively defined affect labels---i.e. the model's output. Third, the identified invariant features can be viewed as a causality-driven feature selection method that avoids capturing spurious correlations in affect modeling. Finally, we validate our approach extensively and across three different games played and annotated by 50 participants each. The validation results suggest that our methodology can efficiently identify and remove inconsistent participants resulting in consistent affective corpora and, thus, yielding more robust models of affect \cite{peters2017elements}. 

\section{Related Work}

This section summarises related literature on techniques that aim to ensure consistent data. It reviews core outlier detection techniques used in machine learning and outlines methods employed to alleviate affect label ambiguities.

\subsection{Outlier Detection in Machine Learning}

Outlier detection is a fundamental data preprocessing task that aims to provide researchers with vital knowledge and assist them in making better decisions about their data. Outlier detection techniques strive to solve the problem of discovering patterns that do not adapt to expected behaviours \cite{ayadi2017outlier}. In typical machine learning applications, deviation from the expected behaviour happens mainly due to noise in the data and failures of sensors that measure a phenomenon. 
Outlier detection techniques can be roughly classified into statistical, clustering, density, and learning-based methods \cite{ranshous2015anomaly}.

Techniques based on statistical analysis use a stochastic distribution to model the data. Data points for which the probability to have been generated by the stochastic distribution model is below a threshold are denoted as outliers \cite{makantasis2017data}. Some of the most commonly used statistical outlier detection methods are based on Gaussian Mixture Models \cite{makantasis2017data,tang2015outlier} and Kernel Density Estimation \cite{papadimitriou2003loci,latecki2007outlier}.

Clustering approaches exploit clustering algorithms to describe data patterns and denote as outliers data points belonging to clusters with few members \cite{zhang2013advancements}. The effectiveness of such approaches is inherently dependent on the effectiveness of the employed clustering algorithm \cite{al2009effective}. For detecting outliers via clustering one may employ partitioning algorithms such as K-Means \cite{georgogiannis2016robust} and CLARANS \cite{ng2002clarans}, hierarchical clustering algorithms \cite{zahn1971graph}, and density-based methods \cite{ankerst1999optics}. Clustering and outlier detection are two different tasks, however. The main drawback of clustering approaches, therefore, is the decision of whether a data point is an outlier or it represents a rare yet important aspect of the data.

Density-based methods consist one of the earliest known approaches to outlier detection, and they often serve as baselines for new algorithms \cite{breunig2000lof,jin2006ranking}. They assume that outliers lie in low-density areas of the data space, while inliers fall into high-density areas \cite{schubert2014local}. Although density-based approaches are entirely data-driven without making any assumption about the distribution of the data, they are sensitive to parameter settings such as determining the neighbourhood size \cite{wang2019progress}.

Finally, learning-based approaches operate in two stages. The first involves training a machine learning model to fit the data, and the second evaluates every data point against the trained model. A data point is denoted as an outlier when the model's output deviates from a ground truth value \cite{dalatu2016comparative}. Due to the recent success of deep learning, learning-based approaches have gained popularity. Deep learning methods, such as generative adversarial networks \cite{creswell2018generative} and autoencoders \cite{protopapadakis2017stacked, eduardo2020robust} are used to learn data representations which then are used with one-class classifiers \cite{oza2018one} to detect outliers.  

\begin{figure*}[!tb]
	\begin{minipage}{1.0\linewidth}
		\centering
		\centerline{\includegraphics[width=1.0\linewidth]{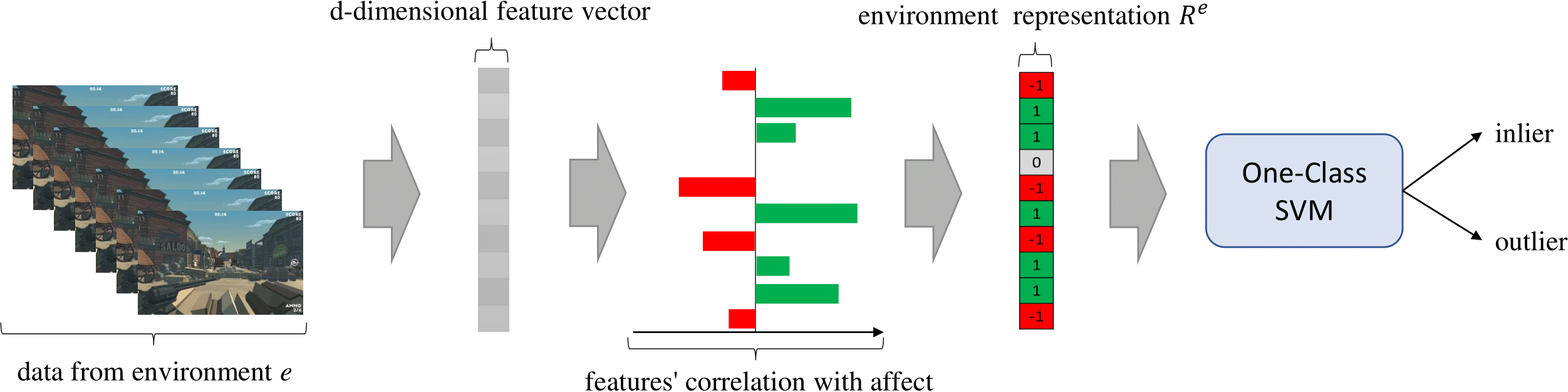}}
	\end{minipage} 
	\caption{The proposed overall methodology for detecting outlier environments.}
	\label{fig:overall}
\end{figure*}

All the above techniques can be successfully applied to typical machine learning problems where the variables are objectively defined. They cannot be straightforwardly applied, however, to AC corpora due to the subjective nature of emotions. In this work, we embrace the subjectivity of affect labels and employ causation tools to formulate an outlier detection method tailored explicitly for affect corpora. 

\subsection{Consistent Labeling of Affect Datasets}

Outlier detection techniques can be used as a data preprocessing step in AC to eliminate data points whose information is corrupted by noise or hardware failures. Beyond that, methodologies that attempt to ensure data consistency in AC primarily focus on the reliability of affect labels. More specifically, these methodologies compare multiple annotations of a particular context (e.g. image, video, sound) as provided by several independent annotators and test for annotation reliability in terms of inter-annotator agreement \cite{artstein2017inter}.

The simplest way to measure agreement is to count the number of items for which the annotators provide identical labels \cite{bayerl2011determines}. This agreement measure is highly expressive and easy to implement; it does not guarantee, however, a reliable annotation process as annotators may agree by mere chance. For this reason annotator agreement---which implies data reliability---is measured alternatively using coefficients from the kappa/alpha family, such as $S$ \cite{bennett1954communications}, $\pi$ \cite{scott1955reliability}. $\kappa$ \cite{cohen1960coefficient,fleiss1971measuring} and $\alpha$ \cite{krippendorff2018content}, which quantify agreement that was attained above the level expected by chance. Cronbach's $\alpha$ has been used for instance in \cite{aljanaki2017developing} to measure inter-annotator agreement and remove data points with unreliable labels. Yannakakis \textit{et al.} \cite{yannakakis2015grounding} employ Krippendorf's $\alpha$ to assess and compare the consistency between ordinal and interval affect labels. In \cite{devillers2006real}, Cohen's $\kappa$ is used for ensuring consensus among annotators. Cronbach's $\alpha$ and Cohen's $\kappa$ are used for assessing inter-annotator agreement in the RECOLA database \cite{ringeval2013introducing}, which has been used for audio/visual emotion recognition challenges.

Extensions for the above coefficients have also been proposed. For example, the study in \cite{bhowmick2008agreement} extends Cohen's $\kappa$ coefficient in scenarios where the annotators have the freedom to provide more than one label to a single data point. In \cite{booth2020fifty} the signed differential agreement metric is proposed for measuring agreement in ordinal and interval-scale annotations, while in \cite{makantasis2019pixels,makantasis2021pixels} dynamic time warping is used to compare continuous annotation traces.


The methodologies above require multiple annotations per data point to measure agreement and provide consistent labels. 
There are AC corpora however in which a) each affect elicitor (e.g. image) is labelled by a single annotator in e.g. a first-person manner and/or b) annotators have annotated different parts of it. The present study proposes a general approach to outlier detection for AC that bypasses the typical third-person multiple-annotation requirement and is applicable to any AC corpus for which different environments can be defined. 


\begin{figure*}[!tb]
	\begin{minipage}{0.33\linewidth}
		\centering
		\centerline{\includegraphics[width=1.0\linewidth]{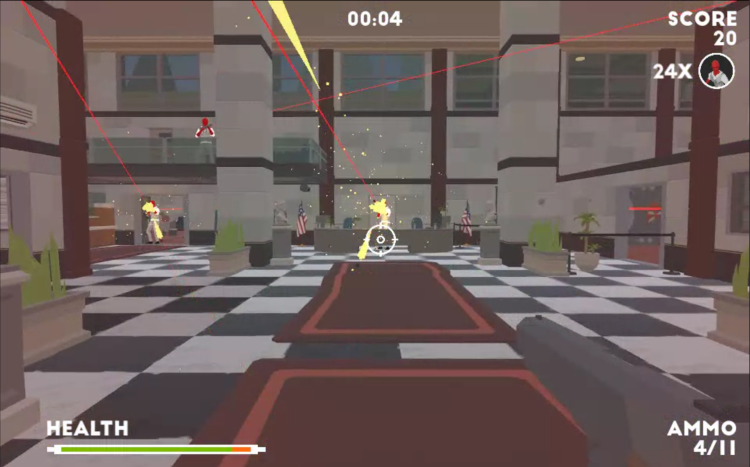}}
	\end{minipage} 
	\begin{minipage}{0.33\linewidth}
		\centering
		\centerline{\includegraphics[width=1.0\linewidth]{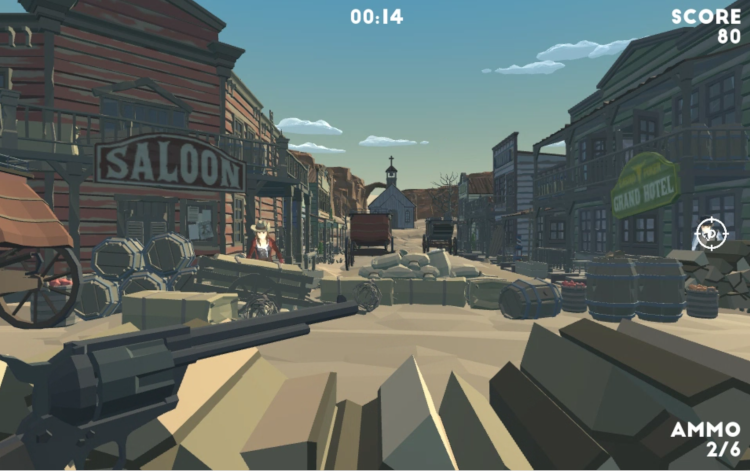}}
	\end{minipage} 
	\begin{minipage}{0.33\linewidth}
		\centering
		\centerline{\includegraphics[width=1.0\linewidth]{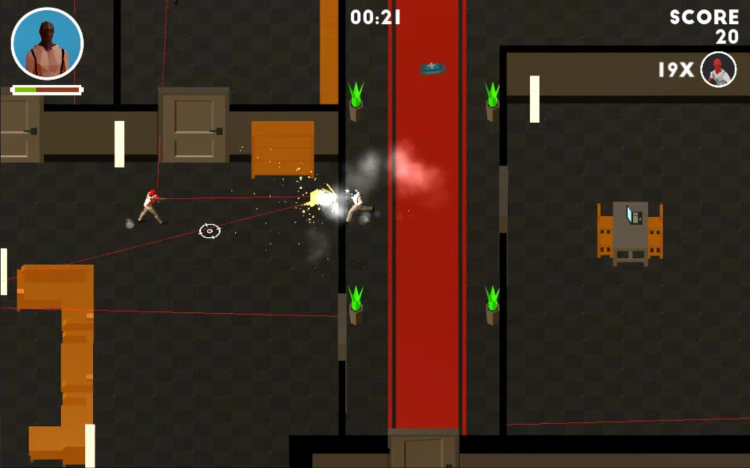}}
	\end{minipage} 
	\caption{The three testbed shooter games of the AGAIN dataset \cite{melhart2022again} used in this study. From left to right: \emph{Heist}, \emph{Shootout}, and \emph{Topdown}.}
	\label{fig:games}
\end{figure*}

\begin{figure}[!tb]
	\begin{minipage}{1.0\linewidth}
		\centering
		\centerline{\includegraphics[width=1.0\linewidth]{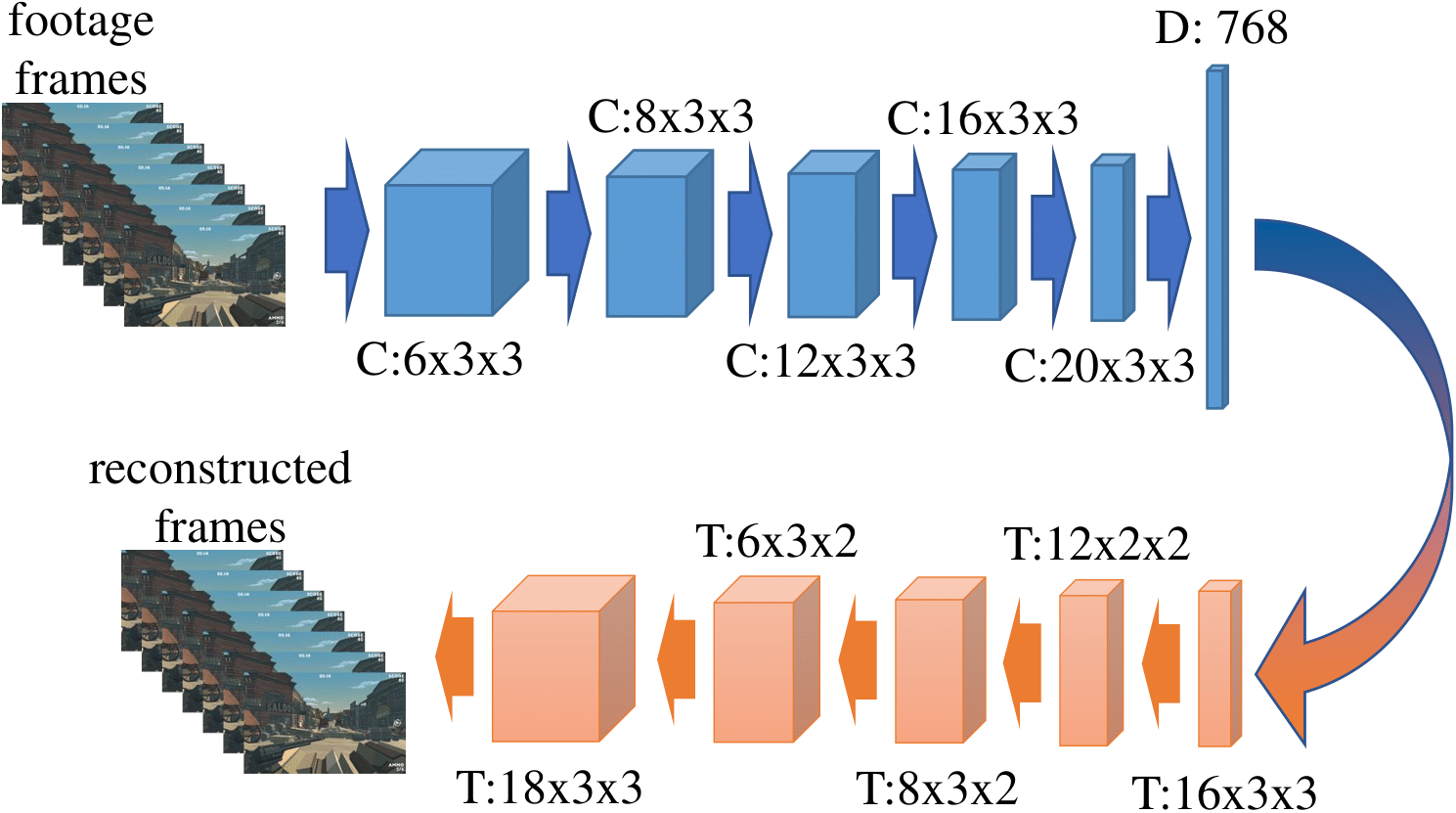}}
	\end{minipage} 
	\caption{The autoencoder architecture employed for constructing high-level representations for the visual information of gameplay time windows. The blue and the orange parts correspond to the encoder and decoder. ``C'', ``D'' and ``T'' stand for convolutional, dense and transpose convolutional layers.}
	\label{fig:autoencoder}
\end{figure}

\section{Invariant Features for Affect Modelling} \label{sec:methodology}


This section presents our proposed methodology for detecting unreliable data points in affective corpora (\ref{sec:methodology_participants}) and identifying invariant features for affect modelling (\ref{sec:methodology_features}). Both unreliable data detection and invariant feature identification are based on the notion of an \textit{environment} that is used in the remainder of this paper. This term is borrowed from causation theory to refer to a mechanism that follows a specific probability distribution for generating data. In this study, we consider different participants as different environments since each participant follows an internal (personal) mechanism for generating data. Data in AC can be annotation traces (or labels) but also affect manifestations, and aspects of the affective interaction. It is important to note that a participant in a third-person annotation scheme is merely an annotator whereas a participant in a first-person annotation scheme can both annotate and generate data that can be annotated.

\subsection{Removing Outliers among Annotators}\label{sec:methodology_participants}

Our approach is based on the assumption that there exist emotion elicitation mechanisms---encoded as stimulus-response rules---that are activated across different environments. Those rules can be represented by features that exhibit a similar relation to affect labels across different environments, i.e. invariant features. Based on the assumption above, we aim to discover those rules and discard data that violates them.  

To define \emph{invariance} we consider the correlation between features and their corresponding affect labels. 
More formally, let us define $\mathcal D^e=\{(X_i^e, Y_i^e)\}_{i=1}^{n^e}$ the data generated and annotated by environment $e \in \mathcal S = \{1,2,\cdots,E\}$. Affect measurements are represented by $d$-dimensional vectors $X_i^e \in \mathbb R^d$ and affect labels by scalars $Y_i^e$. For each environment $e$, we compute the correlation between features $X_i^e$ and labels $Y_i^e$. For continuous labels, the Pearson's correlation coefficient can be used, while for binary labels the point-biserial correlation coefficient suits best. A positive (negative) correlation value between a feature $x$ and affect labels indicates that the linear relation that maps $x$ to labels is monotonically increasing (decreasing). Moreover, an undefined correlation value for a feature $x$ implies that its variance equals zero (the value of $x$ is constant for all data points in $\mathcal D^e$) and, thus, this feature does not carry any information about affect labels.   

Based on the correlation coefficients, we represent the environment $e$ by a $d$-dimensional vector $R^e \in \{-1, 0, 1\}^d$. Specifically, if the correlation between the $i$-th feature and labels is positive, then the $i$-th element of $R^e$ equals 1, if the correlation is negative it equals -1, and if the correlation is undefined or zero it equals 0 (see Fig.~\ref{fig:overall}). This $d$-dimensional environment representation encodes information about the way the different affect measurement features are related to affect labels. Therefore, environments with similar representations will share a large number of invariant features, that is, features that exhibit the same correlation patterns across similar environments. On the contrary, the set of invariant features for environments with dissimilar representations will be small. This implies that modelling affect, using machine learning models which are correlation-based, will be infected by spurious correlations. 

We consider environments whose representation is dissimilar than the representations of the majority of environments as outliers. To detect which environments are outliers we use the One-Class SVM algorithm fed with environments' representations $R^e$ for $e=1, 2,\cdots,E$. One-Class SVM is an unsupervised learning algorithm that models expected data patterns to bring in one class all inliers and discard outliers. Fig.\ref{fig:overall} presents the outlier detection approach, which can be considered end-to-end since it receives as input a set of examples and outputs two clusters of data points: inliers and outliers.


\subsection{Identifying Invariant Features}\label{sec:methodology_features}

A critical byproduct of the methodology presented above is that the representation of the environments encodes information for relationships between features and affect labels that remain invariant across different environments. We exploit this information and use only those invariant features for affect modelling in an attempt to reduce spurious correlations even between similar environments. As mentioned above, the term environment corresponds to mechanisms that generate data by following different probability distributions. For a given affect corpus the number of environments is constant and different environments correspond to different data collection settings, such as different participants/annotators.

To identify invariant features, we follow a simple yet effective approach. Given a set of environments $\mathcal S$, we first discard all affect measurement features for which the correlation with affect labels is 0 or undefined. Then, we count the number of environments $C_{pos}$ ($C_{neg}$) for which a feature is positively (negatively) correlated with affect labels. Finally, we denote a feature as invariant if:
\begin{equation}
\label{eq:lambda}
    \max (C_{pos}, C_{neg}) \geq \lambda|\mathcal S|,
\end{equation}
where $|\mathcal S|$ is the number of environments in $\mathcal S$ and $\lambda \in [0,1]$ is a parameter for balancing between features' invariance and modelling capacity. The value of $\lambda$ is inversely proportional to the number of invariant features. For large values of $\lambda$ the invariant features will not model spurious correlations, but due to their small number they may not exhibit enough modelling power to accurately predict affect. On the contrary, for small values of $\lambda$ the invariance property is disregarded resulting in models of affect infected by spurious correlations. We investigate the impact of parameter $\lambda$ in Section \ref{sec:results}. 

It is important to note that our method is \textit{offline}, in the sense that it requires a batch of data to detect invariant features and inlier/outlier points. The detected invariant features, however, can be used in real-time affect modelling. For outlier samples, these features---like any set of features---will yield poor modelling results. For inlier points, however, the set of invariant features will improve modelling accuracy (see Section \ref{sec:results}).

\section{Dataset} \label{sec:dataset}

This section presents the experimental protocol we designed for collecting the dataset used and the data preprocessing steps.

\subsection{Experimental Protocol}

To test our methodology we selected three games (\emph{Heist}, \emph{Shootout} and \emph{Topdown}) from the AGAIN dataset \cite{melhart2022again} as depicted in Fig.~\ref{fig:games}. These games belong to the shooter genre and require accurate aiming and movement timing. Data was collected from 50 participants that were invited through Amazon's mechanical Turk service. Each participant played a game with a duration of 120 seconds, and then annotated their recorded gameplay footage in terms of arousal using the RankTrace tool implemented in PAGAN \cite{melhart2019pagan}, which allows continuous and unbounded annotations. This play-annotation cycle occurred for all three games. Since the participants had no prior experience in affect annotation, they were presented with an introductory screen that describes arousal as \textit{``the intensity of gameplay no matter whether you like the game or not. High arousal can be a feeling of readiness, tension, excitement or exhilaration. Low arousal can be a feeling of fatigue, boredom, calmness or relaxation''}. Gameplay footage was recorded at 24 frames per second (24Hz) throughout the duration of the game. After play, participants annotated their gameplay video with an arousal trace, providing one arousal value for each of the recorded gameplay frames.

\subsection{Data Preprocessing}
\label{ssec:preprocessing}
To prepare the dataset for testing our methodology, we split the gameplay videos using non overlapping time windows of 3 seconds. Since RankTrace provides continuous and unbounded annotations, we first normalize the arousal annotations to a $[0,1]$ value range in a game session-wise manner. Then, each window is assigned the mean arousal value of its frames. After this data processing step, each window is described by the visual information of 72 frames and a single arousal value. 

To create high-level representations for the visual information of the 3-seconds time windows, we use convolutional autoencoders. Specifically, we  train one convolutional autoencoder with 5 convolutional layers, one dense layer and 5 transposed convolutional layers for each one of the games. For the first 4 convolutional layers we set the stride parameter to 2 and for the fifth layer we set it to 1. For all transposed convolutional layers we set stride to 2 and output padding parameter to 1. We use ReLU as the activation function and we train the autoencoders by minimizing the mean squared reconstruction error using the Adam optimizer. 
After training, we use the encoding part of the autoencoder to construct 768-dimensional representations for each gameplay time window. To reduce the computational cost of training, we convert the RGB frames to grayscale, fix their resolution to $240 \times 150$ pixels, and use frame skipping of 3 frames. As a result, the autoencoders' input consists of 18 grayscale frames concatenated along the channels' dimension. Details about the architecture of the autoencoder are presented in Fig. \ref{fig:autoencoder}.

Having created high-level representations for the visual information of time windows, we form one dataset for each game for preference learning-based affect modelling \cite{yannakakis2018ordinal}. On that basis, we identify all pairs of time windows belonging to the same gameplay. For each pair, we compare the arousal values of their corresponding time windows and we assign label 1 (0) when the arousal value of the first time window is larger (smaller) that that of the second window plus (minus) a threshold $p_t$. 
Finally, we disregard all pairs for which the absolute difference of their arousal values is smaller than $p_t$ to avoid unstable preference learners due to trivial changes in their input. In this study, we set $p_t$ to $0.15$ which is slightly stricter that the threshold proposed in \cite{melhart2022again}. We represent every pair by a 768-dimensional latent vector, which corresponds to the representations difference of the first and second time window \cite{camilleri2019pyplt,joachims2002optimizing}. Following the pairwise data transformation phase above, we end up with 50,254, 49,186 and 55,166 data points for Topdown, Heist and Shootout games, respectively. 

\begin{figure}[t]
	\begin{minipage}{1.0\linewidth}
		\centering
		\centerline{\includegraphics[width=1.0\linewidth]{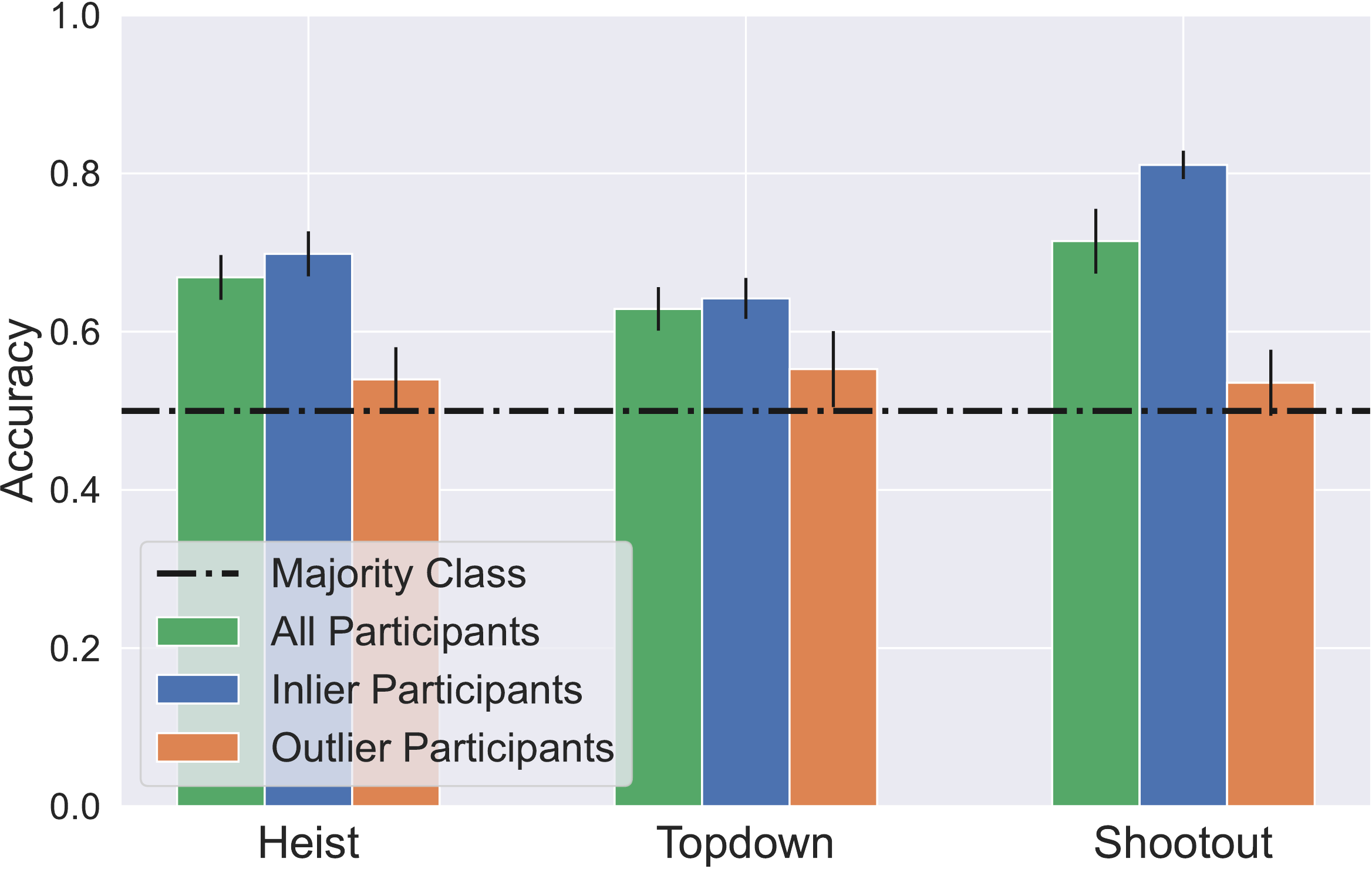}}
	\end{minipage} 
	\caption{Performance of the logistic regression affect model using all participants, the identified outlier participants, and the inlier participants.} 
	\label{fig:barchart}
\end{figure}

\section{Results} \label{sec:results}

In this section we present the experimental validation of our proposed methodology. We first present the results of the proposed outlier detection method (Section \ref{sec:detection}) and we then present our affect modelling experiments using invariant features (Section \ref{sec:modelling}). For all performed experiments in this paper we evaluate the performance of our methodology in terms of preference prediction accuracy. Preference learning \cite{furnkranz2011preference} can be seen as a binary classification task---preferred vs. non-preferred---where the two classes are perfectly balanced. Therefore, the prediction accuracy of a classifier that predicts the majority class in the training set is $0.5$; this classifier serves as baseline. In this study we choose to use linear classifiers to be able to evaluate the performance of the proposed outlier detection method and affect modelling using invariant features without fiddling with complex nonlinear machine learning models, which add extra parameters for investigation. 

Following the above preprocessing steps described in Section \ref{ssec:preprocessing}, we apply the outlier detection techniques on the time windows representations produced by our autoencoder. Moreover, since our affect labels are binary we use point-biserial correlation for creating the participants' (environments) representations fed to the One-Class SVM. Finally, invariant features correspond to subsets of the 768-dimensional representations produced by the autoencoder. 

\subsection{Detecting Outlier Annotators} \label{sec:detection}

To investigate the effectiveness of the proposed outlier detection method, we first remove the outliers by applying the method presented in Section \ref{sec:methodology_participants}, and then train three classifiers with the 768-dimensional representations obtained from our autoencoders. The first classifier uses data from all available participants without discriminating between inliers and outliers. The second and the third classifiers were trained using data from inlier and outlier participants, respectively. This way, we are able to evaluate modelling accuracy before and after outlier removal. To evaluate the performance of the classifiers we use the demanding leave-one-participant-out cross validation scheme and report classifiers' preference prediction accuracy and $95\%$ confidence intervals. Note that the classifiers are evaluated on different testing data. By using, however, classifiers of the same capacity and treating the training/testing sets as random variables we are able to objectively evaluate the quality of the outliers’ detection result (see \cite{langford2005tutorial} Theorem 3.3).

After the application of the proposed outlier detection technique, 16 participants are denoted as outliers for the Heist and Shootout games and 8 for the Topdown game. Fig. \ref{fig:barchart} presents the preference prediction accuracy and the $95\%$ confidence intervals for the three classifiers mentioned above. The classifier that uses only inliers performs better (in the case of the Shootout game significantly better) than the classifier that uses data from all participants, and significantly better than the classifier trained only on outliers in all games. These findings suggest that the data from inlier participants is indeed consistent. Moreover, we can observe that the performance of the classifier that uses data from the set of outliers is slightly better than the performance of the baseline classifier.

\begin{table}[t]
	\centering
	\caption{Classifiers' accuracy and 95\% confidence intervals trained on outliers set, randomly selected inliers sets and randomly selected participants from the pool of 50 participants.}
	\newcolumntype{L}[1]{>{\hsize=#1\hsize\raggedright\arraybackslash}X}%
	\newcolumntype{C}[1]{>{\hsize=#1\hsize\centering\arraybackslash}X}%
	\label{table:1}
	
	\begin{tabularx}{0.98\linewidth}{L{8.3}C{5.6}C{5.6}C{5.6}}
		\hline \hline 
		& Heist & Topdown & Shootout \\ \hline
		
		Outliers set & 0.54 $\pm$ 0.041  & 0.55 $\pm$ 0.048 & 0.53 $\pm$ 0.042\\ \hline
		Random inliers sets  & 0.67 $\pm$ 0.010  & 0.62 $\pm$ 0.015  & 0.76 $\pm$ 0.006\\ \hline
        Random part. & 0.60 $\pm$ 0.009  & 0.60 $\pm$ 0.013  & 0.65 $\pm$ 0.016\\ \hline
	    \hline
	\end{tabularx}
\end{table}

\begin{figure*}[t]
	\begin{minipage}{0.33\linewidth}
		\centering
		\centerline{\includegraphics[width=1.0\linewidth]{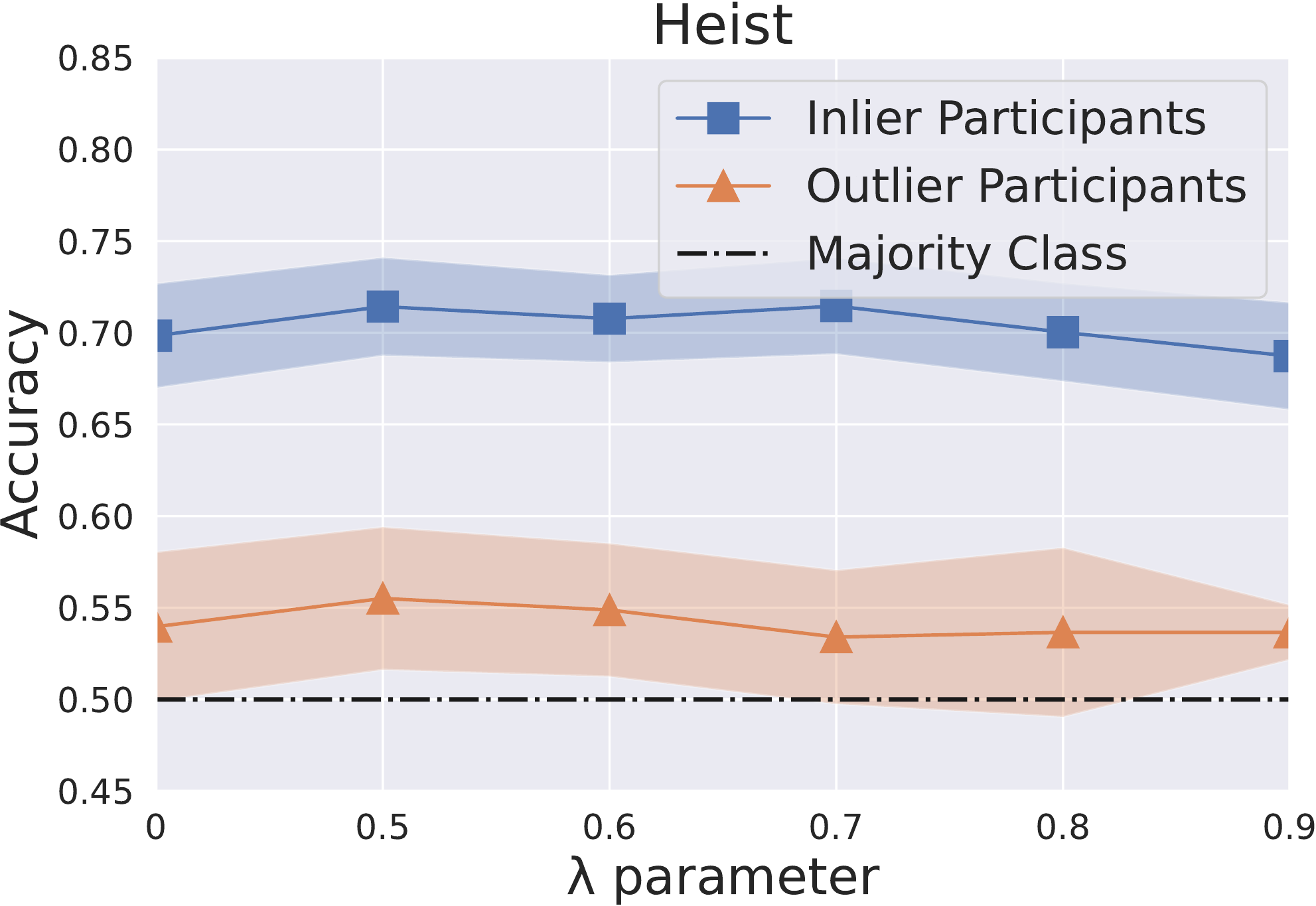}}
	\end{minipage} 
	\begin{minipage}{0.33\linewidth}
		\centering
		\centerline{\includegraphics[width=1.0\linewidth]{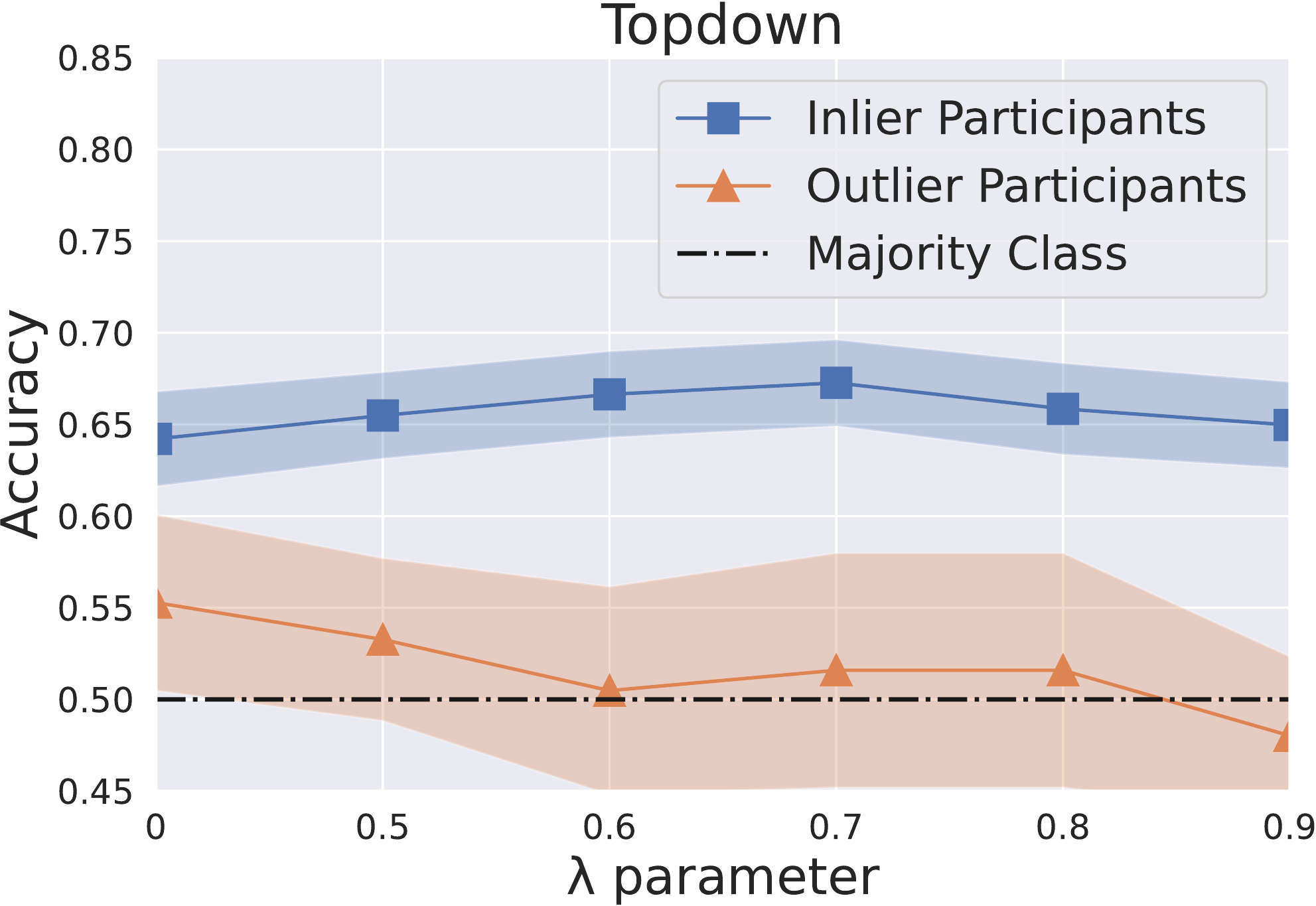}}
	\end{minipage} 
	\begin{minipage}{0.33\linewidth}
		\centering
		\centerline{\includegraphics[width=1.0\linewidth]{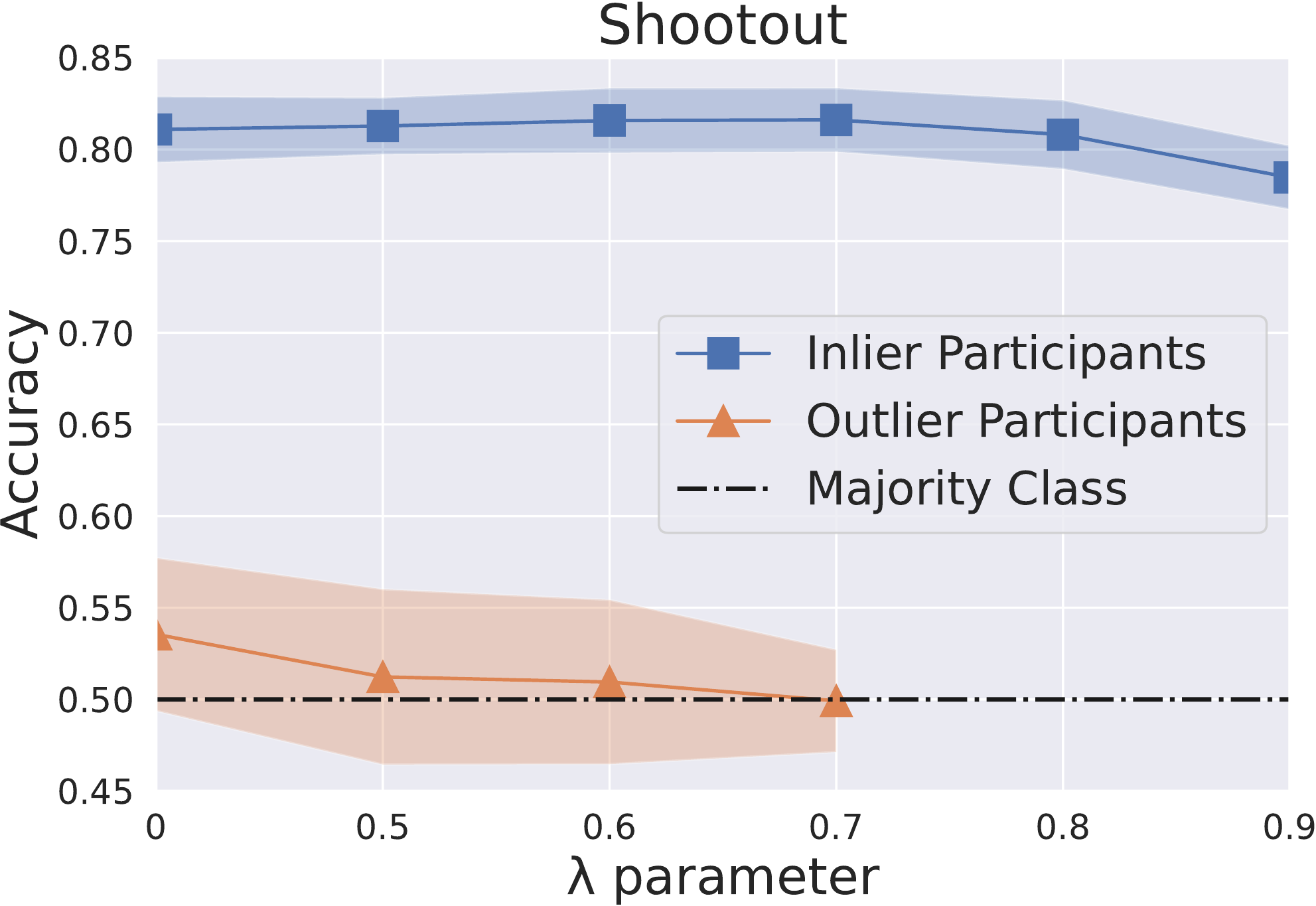}}
	\end{minipage}
	
	\vspace{0.05in}
	
	\begin{minipage}{0.33\linewidth}
		\centering
		\centerline{\includegraphics[width=1.0\linewidth]{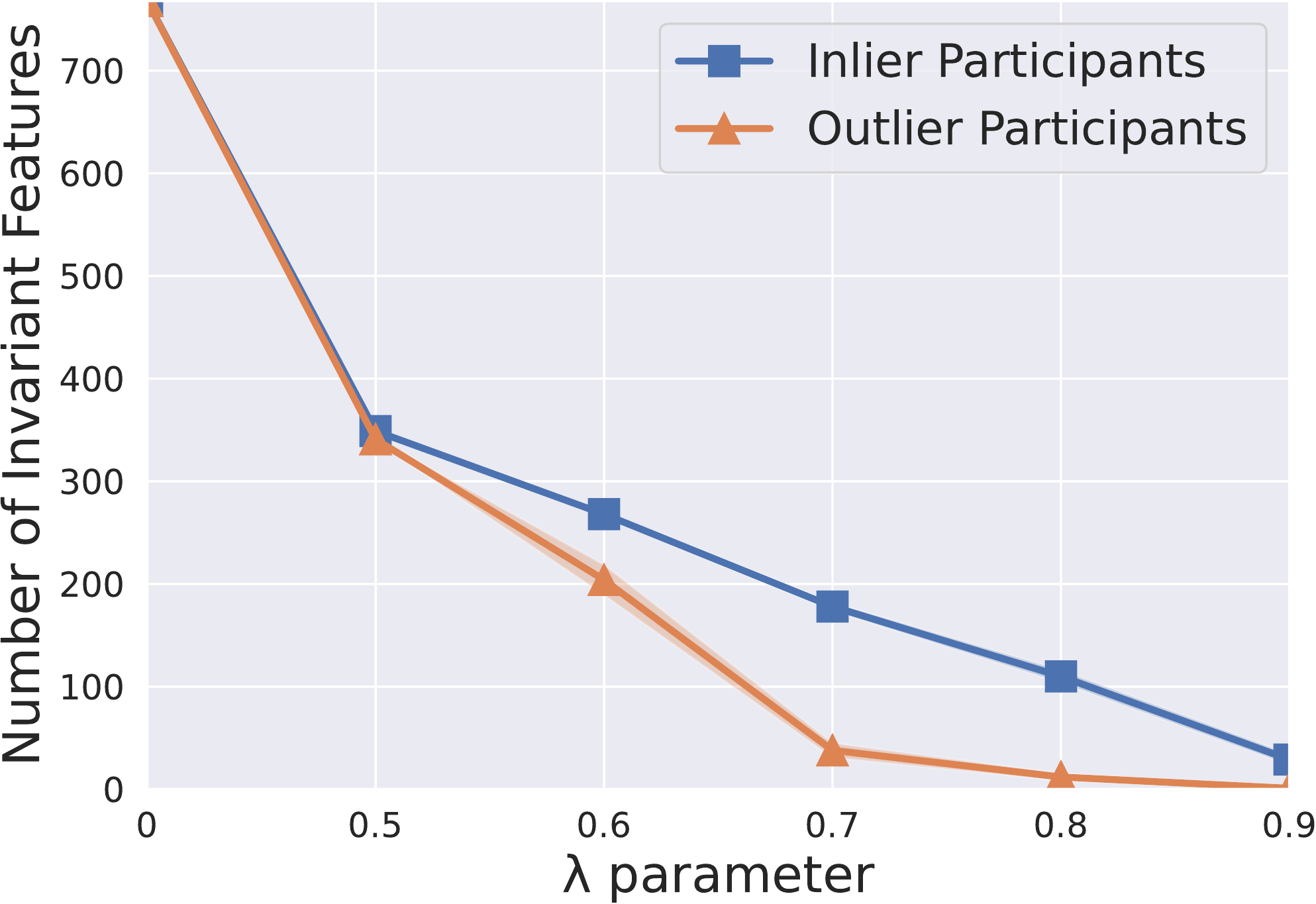}}
	\end{minipage} 
	\begin{minipage}{0.33\linewidth}
		\centering
		\centerline{\includegraphics[width=1.0\linewidth]{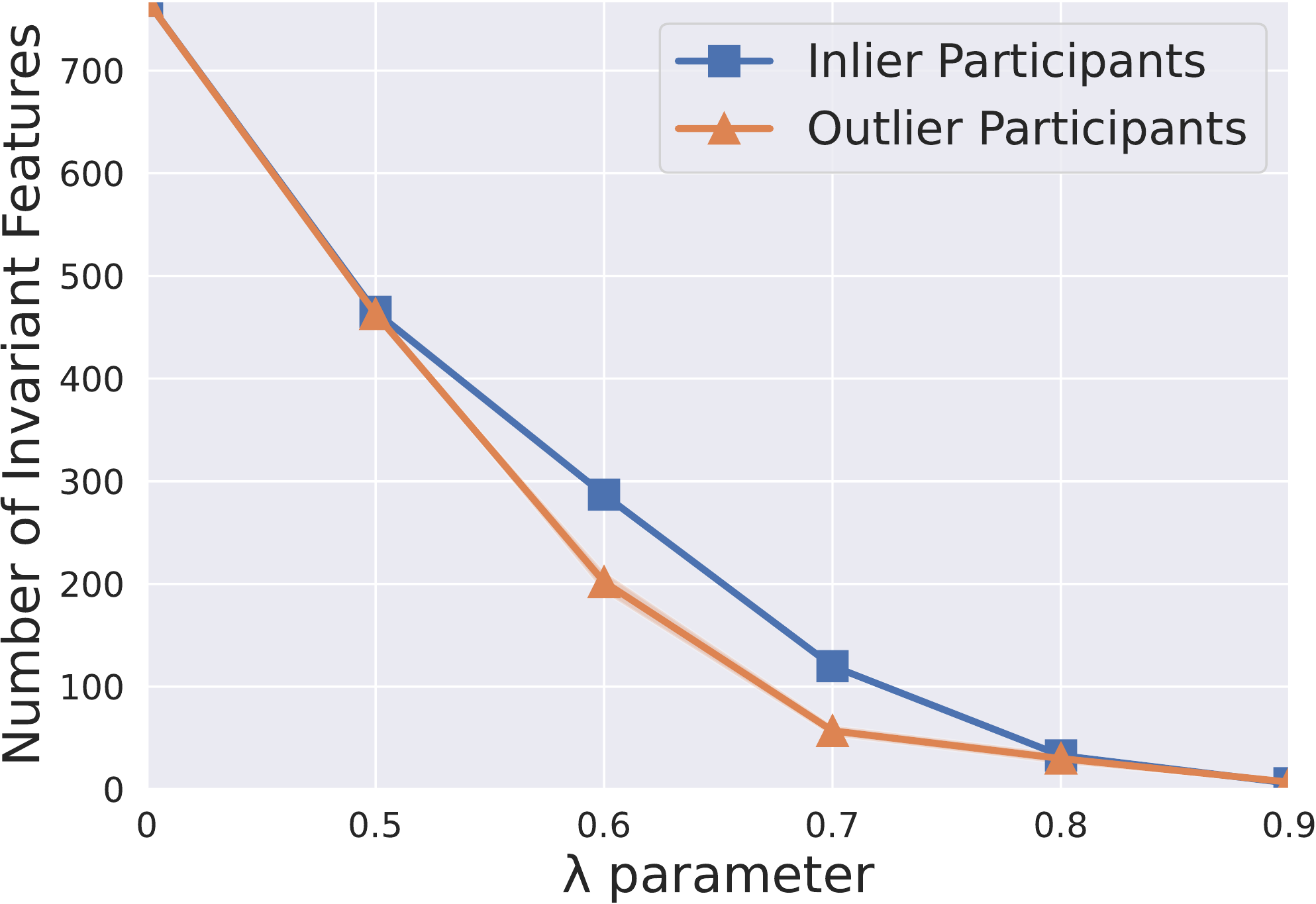}}
	\end{minipage} 
	\begin{minipage}{0.33\linewidth}
		\centering
		\centerline{\includegraphics[width=1.0\linewidth]{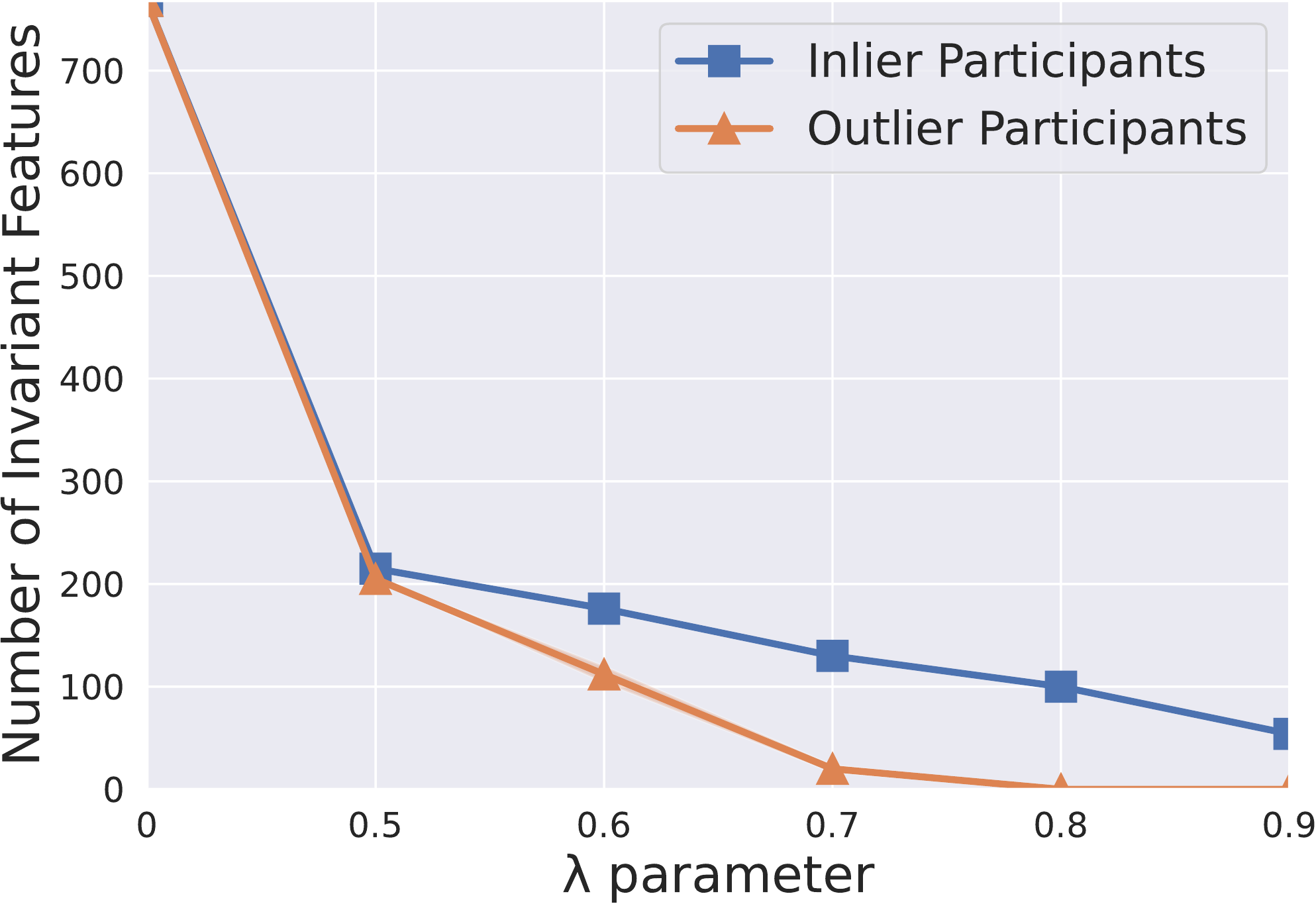}}
	\end{minipage}
	\caption{Affect modelling accuracy and $95\%$ confidence intervals (top row), and number of features used (bottom row) with respect to parameter $\lambda$.}
	\label{fig:lambda}
\end{figure*}

To ensure that this drop in performance is not due to the size of the dataset (i.e. the outliers dataset is smaller than the inliers dataset), we run 30 independent experiments using data from randomly selected inliers such that the number of inliers is equal to the number of outliers; i.e. 16 randomly selected inliers for the Heist and Shootout games and 8 inliers for the Topdown game. These classifiers yield accuracies of $0.67$, $0.62$ and $0.76$ for the Heist, Topdown and Shootout games, respectively (see Table \ref{table:1}). These accuracies show a relative improvement of $24\%$, $13\%$ and $43\%$ compared to corresponding classifiers trained on the set of outliers. This indicates that the participants belonging to the set of outliers exhibit different patterns for playing and annotating the data. Due to this difference, outlier data are not consistent, resulting in affect models that are not able to capture the emotion dynamics. 
To benchmark the inlier detection algorithm against random chance,  we also ran 30 independent experiments using data from randomly selected participants from the pool of 50 participants such that their number equals the number of outliers. Then, we train a classifier using their data (which may come both from the inlier and the outlier sets). These classifiers achieve accuracies of $0.60$ for the Heist and Topdown games and $0.65$ for the Shootout game, which have a relative drop of $11\%$, $9\%$ and $22\%$ compared to the accuracies of classifiers trained only on inliers sets with the same cardinality (see Table \ref{table:1}). We can, therefore, argue that the outcome of the proposed outlier detection technique is not obtained by chance.
%

Based on all aforementioned results of this investigation, we can conclude that exploiting correlation patterns across different environments is an effective way to detect outliers.

\subsection{Invariant Affect Modelling} \label{sec:modelling} 

In this section we investigate the impact of invariant features on arousal preference modelling. For that purpose we train classifiers using the feature sets obtained for different values of parameter $\lambda$ in Eq.~\eqref{eq:lambda}. We train classifiers for inlier and outlier participants and report their average accuracy and $95\%$ confidence intervals following, once more, the leave-one-participant-out cross validation scheme. 

The results of this investigation are summarized in Fig.~\ref{fig:lambda}. For all games the classifiers trained on inliers achieve the maximum accuracy for $\lambda = 0.7$. Specifically, these classifiers  yield accuracy values of $0.71$, $0.67$ and $0.82$ for the Heist, Topdown and Shootout games, respectively. By setting the parameter $\lambda$ to 0.7 the number of invariant features for the Heist game is 178, while for the Topdown and Shootout games are 120 and 130, respectively. We observe that our approach can reduce the number of features used for modelling arousal (initially 768) by identifying those features that are invariant across inlier participants and yield high modelling capacity. 

The results obtained on the outliers set, however, do not seem to follow a specific pattern. Irrespective of the $\lambda$ value, the average accuracy of the classifier is around 0.5, on par with random guessing. This suggests that the set of invariant features among the outliers that belong to the training set cannot be used for modelling the preferences of the outlier participant belonging to the test set. This highlights further the effectiveness of our proposed outlier detection technique; evidently the participants denoted as outliers all appear to follow inconsistent patterns between gameplaying (i.e. experiencing the context) and providing arousal annotations for their game.  

By examining the bottom row of Fig.~\ref{fig:lambda}, we observe that the number of invariant features for the set of inliers is larger than the number of invariant features for the outliers. This observation is in alignment with the expected behaviour of the outlier detection scheme. The set of inliers comprises of data that are consistent: i.e. participants denoted as inliers follow similar patterns for playing and annotating the three games. Therefore, the probability the features of inlier data to exhibit similar correlation patterns with arousal labels is higher compared to outlier data. Due to this, the number of invariant features for the inlier data is larger than the corresponding number for the outlier participants. Interestingly, for the Shootout game there are no invariant features for the set of outliers for $\lambda$ values larger than 0.7. We also observe that the accuracy of the classifier on the set of inlier participants drops slightly for $\lambda$ values larger than $0.7$. As discussed in Section \ref{sec:methodology} this behaviour is also expected. Parameter $\lambda$ is used for balancing features' invariance and modelling capacity. For $\lambda$ values larger than $0.7$ the selected features are invariant across the vast majority of participants; however, due their their small number they cannot accurately capture the arousal dynamics.   

Based on the results above, we can conclude that employing invariant features---borrowed by causation theory---is an effective way towards deriving robust models of affect. 
Finally, parameter $\lambda$ seems to affect the robustness of the models and, thus, it should be treated as a hyperparameter which should be appropriately set according to the affective corpus at hand.

\section{Discussion and Conclusions}

In this paper we exploit invariant features, borrowed from causation theory, to derive an outlier detection method tailored for affective corpora and to build robust models of affect. Our proposed methodology is based on the assumption that there exist features of the experienced context (i.e. the affect elicitor) and the manifested affect that exhibit consistent relationships with annotated affect across different environments; environments are determined by different participants in this study. We identify those features via a correlation analysis and we name them \emph{invariant}. Based on the correlation patterns, we first create data representations for detecting outliers and then we use the set of invariant features to derive robust models of affect. We test our methodology within the domain of digital games by using data from three testbed games played and annotated in terms of arousal by 50 participants. The experimental findings suggest that our methodology can successfully denote as outliers those participants that present playing and annotation behaviour that is different than the majority of participants. Moreover, by using invariant features we are able to better capture arousal dynamics and build more accurate models of affect. 

The presented methodology can be applied to any affective corpus comprising data from multiple environments. Its potential limitation, however, stems from the fact that the current implementation exploits linear correlation metrics to discover invariance. We thus need to further investigate the degree to which this linearity restricts the efficiency of our method. Towards this direction, we aim to employ non-linear correlation metrics and even learning-based approaches \cite{arjovsky2019invariant} for discovering invariant features. Moreover, besides linear classifiers, we aim to investigate the behaviour of invariant features in conjunction with non-linear models. In addition, this study focuses only on visual information. By using, however, general purpose representations of visual content, we do not restrict our method to specific tasks. Nevertheless, we aim to further validate its efficiency on data from diverse AC corpora that employ different and multiple modalities of affect. Finally, we aim to extend this approach by deriving several clusters (not only inliers/outliers) of data points that share common characteristics to avoid removing examples which makes the data less representative. 
Based on the clustering results we expect to be able to build cluster-wise models of affect based on invariant features.

To conclude, to the best of our knowledge, this is the first attempt towards integrating methods from causation theory to affect modelling. Deriving models of affect that express causation is extremely important for the field. We believe that causation can provide valuable tools for disentangling complex relations between the context, its affect manifestations and corresponding affect labels and thereby identifying general cause-effect relationships or underlying rules of affect elicitation.
At this point we should mention that some studies (e.g. \cite{mittal2021affect2mm}) propose the use of Granger causality \cite{friston2014granger} in affect modelling. This type of causality, however, tests the degree to which temporal information about affect measurements can forecast the experienced emotions. Therefore, Granger causality cannot discover cause-effect relations and, in general, it is not considered as a tool derived from causation theory. 
The proposed methodology has direct applications to any affect modelling task that employs data from multiple environments. In stark contrast to previously proposed methods for cleaning AC corpora, 
our method does not require multiple annotations per data point significantly reducing the data annotation cost.

\bibliographystyle{IEEEtran}

\end{document}